\documentclass[conference]{IEEEtran}
\usepackage{verbatim}

\usepackage{cite}
\ifCLASSINFOpdf
   \usepackage[pdftex]{graphicx}
\else
\fi
\usepackage{subcaption}
\usepackage[cmex10]{amsmath}
\usepackage[]{algorithm2e}

\usepackage{mdwmath}
\usepackage{mdwtab}
\usepackage[caption=false,font=footnotesize]{subfig}
\usepackage{caption}
\usepackage{subcaption}
\usepackage{subfig} 
\newcommand\lword[1]{\leavevmode\nobreak\hskip0pt plus\linewidth\penalty50\hskip0pt plus-\linewidth\nobreak{#1}}

\begin{document}
%
% paper title
% can use linebreaks \\ within to get better formatting as desired
\title{From Word Embeddings to Item Recommendation}

% author names and affiliations
% use a multiple column layout for up to three different
% affiliations

\author{
\IEEEauthorblockN{Makbule Gulcin Ozsoy}
\IEEEauthorblockA{Department of Computer Engineering \\ Middle East Technical University \\Ankara, Turkey\\
Email: makbule.ozsoy@ceng.metu.edu.tr}

%\and
%\IEEEauthorblockN{Homer Simpson}
%\IEEEauthorblockA{Twentieth Century Fox\\
%Springfield, USA\\
%Email: homer@thesimpsons.com}
%\and
%\IEEEauthorblockN{James Kirk\\ and Montgomery Scott}
%\IEEEauthorblockA{Starfleet Academy\\
%San Francisco, California 96678-2391\\
%Telephone: (800) 555--1212\\
%Fax: (888) 555--1212}
}

% conference papers do not typically use \thanks and this command
% is locked out in conference mode. If really needed, such as for
% the acknowledgment of grants, issue a \IEEEoverridecommandlockouts
% after \documentclass

% for over three affiliations, or if they all won't fit within the width
% of the page, use this alternative format:
% 
%\author{\IEEEauthorblockN{Michael Shell\IEEEauthorrefmark{1},
%Homer Simpson\IEEEauthorrefmark{2},
%James Kirk\IEEEauthorrefmark{3}, 
%Montgomery Scott\IEEEauthorrefmark{3} and
%Eldon Tyrell\IEEEauthorrefmark{4}}
%\IEEEauthorblockA{\IEEEauthorrefmark{1}School of Electrical and Computer Engineering\\
%Georgia Institute of Technology,
%Atlanta, Georgia 30332--0250\\ Email: see http://www.michaelshell.org/contact.html}
%\IEEEauthorblockA{\IEEEauthorrefmark{2}Twentieth Century Fox, Springfield, USA\\
%Email: homer@thesimpsons.com}
%\IEEEauthorblockA{\IEEEauthorrefmark{3}Starfleet Academy, San Francisco, California 96678-2391\\
%Telephone: (800) 555--1212, Fax: (888) 555--1212}
%\IEEEauthorblockA{\IEEEauthorrefmark{4}Tyrell Inc., 123 Replicant Street, Los Angeles, California 90210--4321}}

% use for special paper notices
%\IEEEspecialpapernotice{(Invited Paper)}

% make the title area
\maketitle

\begin{abstract}
Social network platforms can use the data produced by their users to serve them better. One of the services these platforms provide is recommendation service. Recommendation systems can predict the future preferences of users using their past preferences. In the recommendation systems literature there are various techniques, such as neighborhood based methods, machine-learning based methods and matrix-factorization based methods. In this work, a set of well known methods from natural language processing domain, namely Word2Vec, is applied to recommendation systems domain. Unlike previous works that use Word2Vec for recommendation, this work uses non-textual features, the check-ins, and it recommends venues to visit/check-in to the target users. For the experiments, a Foursquare check-in dataset is used. The results show that use of continuous vector space representations of items modeled by techniques of Word2Vec is promising for making recommendations.


\end{abstract}

% IEEEtran.cls defaults to using nonbold math in the Abstract.
% This preserves the distinction between vectors and scalars. However,
% if the conference you are submitting to favors bold math in the abstract,
% then you can use LaTeX's standard command \boldmath at the very start
% of the abstract to achieve this. Many IEEE journals/conferences frown on
% math in the abstract anyway.

% no keywords

% For peer review papers, you can put extra information on the cover
% page as needed:
% \ifCLASSOPTIONpeerreview
% \begin{center} \bfseries EDICS Category: 3-BBND \end{center}
% \fi
%
% For peerreview papers, this IEEEtran command inserts a page break and
% creates the second title. It will be ignored for other modes.
\IEEEpeerreviewmaketitle

\section{Introduction}\label{intro}
Social network platforms (e.g. Twitter, Facebook, Foursquare) have many active users who produce vast amount of information by interacting with each other and with items/services the platform provide. For example, up to April 2016, 320 million monthly active users use Twitter, more than 50 million users use Foursquare and 1.04 billion daily active users use Facebook. These platforms are able to archive and use the produced information to better serve their users. One of the services that most of the social network platforms provide is recommendation service. 

Recommendation systems predict the future preferences of users' based on their previous interactions with the items. For example, information on previous check-ins of users can be used to make recommendations on future check-ins. The vast amount of information produced by the users is used by several different methods to make recommendations, e.g. neighborhood based methods, machine-learning based methods and matrix-factorization based methods. Recently, matrix factorization (MF) methods gained more attention by researchers, as these methods can efficiently deal with large datasets by using low-rank approximation of input data \cite{Ma:2011:RSS:1935826.1935877}. 

Similar to matrix factorization methods, word embedding methods learn low-dimensional vector space representation of input elements. They are used to learn linguistic regularities and semantic information from large text datasets and they are gaining more attention especially in natural language processing and text mining fields \cite{MustoSGL15}. In this work, we aim to recommend next venues to visit/check-in by adopting skip-gram and continuous bag of words (CBOW) word embedding techniques proposed in Word2Vec framework \cite{MikolovSCCD13}.

Efficiency of using text processing techniques in recommendation systems is already exemplified in some of the previous works in the literature (\hspace{1sp}\cite{GaoTL12}, \cite{ShinCL14}, \cite{MustoSGL15}). \cite{GaoTL12} is one of the state-of-the-art methods for venue recommendation on Location Based Social Networks (LBSNs) and employs a language model based method. \cite{ShinCL14} aims to make recommendation to users about which blog to follow. It uses Word2Vec to model a word based feature, i.e. tags. \cite{MustoSGL15} employs three different word embedding techniques, one of which is Word2Vec, to make recommendation on MovieLens and DBbook datasets. It uses textual data collected from Wikipedia about the items. Unlike the previous works that use Word2Vec for recommendation, in this work a non-textual feature, namely the past check-ins of the users, is used to make recommendations.

The aim of this work is to apply a set of well known methods from natural language processing domain, namely methods from Word2Vec framework, to recommendation systems domain and show that the performance is comparable to well-known methods in recommendation systems. The contributions of this work are as follow:
\begin{itemize}
\item Methods from natural language processing domain, \lword{Word2Vec's} skip-gram and continuous bag of words (CBOW) techniques, are employed to make recommendations on Location Based Social Networks (LBSNs).
\item Several different techniques inspired from well-known recommendation methods, namely content based and collaborative filtering based methods, are combined with Word2Vec's techniques. %This shows that it is possible to integrate methods from different domains. 
\item Unlike the previous works that use Word2Vec for recommendation (\hspace{1sp}\cite{ShinCL14}, \cite{MustoSGL15}), a non-textual feature, namely the past check-ins of the users, is used to make recommendations. 
\item For the evaluation a Foursquare check-in dataset, which is already used in previous works (\hspace{1sp}\cite{GaoTL12}, \cite{Ozsoy14}) is employed. Also comparisons to methods from the recommendation systems literature are presented.
\end{itemize}

The rest of the paper is structured as follows: Information on the related work is given in the Section \ref{relWork}. Word2Vec and the proposed methods are explained in the Section \ref{appDeepRec}. The experimental results and the comparisons are given in the Section \ref{eval}. The paper is concluded in the Section \ref{conclusion}.

\section{Related Work}\label{relWork}
Recommendation systems make recommendation of items by estimating the preferences of users (\hspace{1sp}\cite{MassaA07}, \cite{TavakolifardA12}). In the literature there are three base recommendation approaches: Content based, collaborative filtering and hybrid approaches. Content based approach uses features of items and users, calculates their similarities and use these similarities to make recommendations. Collaborative filtering approach uses past preferences of users to decide which items to recommend. Hybrid methods combine these approaches to make recommendations.

%The data to be processed by a recommendation system has basically three elements, which are \emph{user}, \emph{item} and \emph{rating}. In most of the algorithms, these elements are represented by a matrix or a graph. 

Beside the above-mentioned methods, there are various methods to make recommendation, e.g. by neighborhood based methods, machine-learning based methods and matrix-factorization based methods. Recently, the matrix factorization based methods gained more attention of recommendation systems researchers. These methods use low-rank approximation of input data and can handle large volume of data \cite{Ma:2011:RSS:1935826.1935877}. In \cite{KorenBV09}, it is stated that matrix factorization can represent the items and the users as vectors, where high correlation between vectors leads to recommendation. Also, in the same work it is stated that these methods have good scalability, high accuracy and flexibility. Some example works that use the matrix factorization for recommendation belong to Sarwar et al. \cite{sarwar2000}, Ma et al. \cite{Ma:2008:SSR:1458082.1458205}, Zheng et al. \cite{Zheng:2010:CLA:1772690.1772795}, Liu et al. \cite{Liu:2013:SSN:2488388.2488457}, Yu et al. \cite{hfy12a}, Cheng et al. \cite{Cheng:2013:YLG:2540128.2540504} and Lian et al.\cite{Lian:2014:GJG:2623330.2623638}. Among these works \cite{Zheng:2010:CLA:1772690.1772795} and \cite{Cheng:2013:YLG:2540128.2540504} have similar purpose as ours and they make location/activity recommendations to the target users. However none of these methods employ Word2Vec for this purpose.

Similar to matrix factorization methods, word embedding methods from natural language processing field learn low-dimensional vector space representation of input elements. The word embeddings learn linguistic regularities and semantic information from the input text datasets and represent the the meaning of the words by a vector representation (\hspace{1sp}\cite{MustoSGL15},  \cite{AroraLLMR15}). In \cite{AroraLLMR15} it is stated that word embeddings can be learned by Latent Semantic Analysis (LSA), topic models and matrix factorization techniques. Techniques defined in Word2Vec \cite{MikolovSCCD13}, namely skip-gram and CBOW, are commonly used in the literature to represent the word vectors. 

Some of the recommendation methods (\hspace{1sp}\cite{ShinCL14}, \cite{MustoSGL15}) use techniques from Word2Vec to represent their text based features. \cite{ShinCL14} aims to make recommendation to users about which Tumblr blogs to follow. In that work inductive matrix completion (IMC) method is used for recommendation. That method uses side features (i.e. likes, re-blogs and tags) as well as past preferences of users. It does not directly use techniques from natural language processing, but employ Word2Vec to compute vector representation of tags; which are word based features. \cite{MustoSGL15} empirically evaluates three word embedding techniques, namely Latent Semantic Indexing, Random Indexing and Word2Vec, to make recommendation. They evaluate their proposed method on MovieLens and DBbook datasets. They mapped the items in the datasets to textual contents using Wikipedia and used the textual contents for making recommendation. Another recommendation method that uses techniques from natural language processing is Socio-Historical method proposed in \cite{GaoTL12}. It is one of the state-of-the-art methods for venue recommendation on LBSNs. Observing the similarities in text mining and social network datasets, it employs language models approach from natural language processing to make venue recommendations. It models either users' historical preferences or their social interactions or both together. 

Techniques in Word2Vec are generally considered as deep learning technique. There are few other methods that employ deep learning to make recommendations, e.g. \cite{SalakhutdinovMH07}, \cite{GeorgievN13} and \cite{WangWY14}. \cite{SalakhutdinovMH07} uses Restricted Boltzmann Machines (RBM's) to make movie recommendations. It models correlation among item ratings. \cite{GeorgievN13} extends \cite{SalakhutdinovMH07} by modelling both user-user and item-item correlations. \cite{WangWY14} proposes a hierarchical \lword{Bayesian} model that learns models on both content information on items and past preferences of users.

In this work, Word2Vec's skip-gram and CBOW techniques are employed to recommend check-in locations to the target users. Unlike the previous works that use Word2Vec for recommendation (\hspace{1sp}\cite{ShinCL14}, \cite{MustoSGL15}), a non-textual features, namely the past check-ins of the users, is used.  

\section{Recommendation using Multiple Data Sources}\label{appDeepRec}
The aim in this work is to list the top-k venues/locations (e.g. restaurant, cafe) that the target user will visit/check-in in the future. For this purpose techniques from Word2Vec toolbox, namely skip-gram and CBOW, are used. In this section, a brief information on the techniques in Word2Vec toolbox and explanation on how they are used for venue recommendation is presented.

\begin{figure}
\centering
\includegraphics[width=0.95\linewidth]{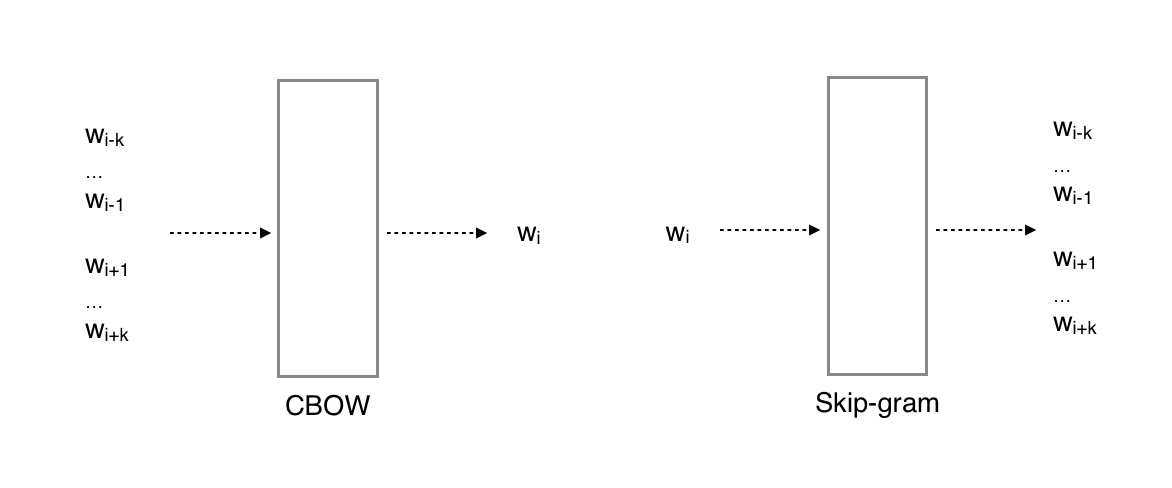}
\caption{Word2Vec techniques}
\label{wordvecImg}
\end{figure}

Word2Vec is a group of models which is introduced by Mikolov et al. (\hspace{1sp}\cite{MikolovSCCD13}, \cite{MikolovCoRR13}). It contains two different techniques, namely skip-gram and CBOW, which produce word embeddings, i.e. distributed word representations. The word embeddings represent the words in a low dimensional continuous space and carry the semantic and syntactic information of words \cite{LiXTJZC15}. While the CBOW technique uses the words around the current word to predict the current word, the skip-gram technique does the vice-versa, such that it uses the current word to predict the words around the current word (Figure \ref{wordvecImg}). In both of the techniques, bag-of-words representation is used, i.e. order of the words in the input does not affect the result. We used both of the techniques to model the data and to make recommendations.

The proposed recommendation method is composed of the following steps: First the input data is modeled using the techniques from Word2Vec. Then the output model is used to execute the recommendation process.

\subsection{Modeling the data using the techniques from Word2Vec} 
We used the Word2Vec techniques implemented in the gensim toolbox \footnote{https://radimrehurek.com/gensim/models/word2vec.html}. This implementation accepts a list of sentences which are themselves are a list of words. These words are used to create the internal dictionary which holds the words and their frequencies. Afterwards the model is trained using the input data and the dictionary. The output of the technique is continuous vector representation of words, which can be used as features by different applications\cite{Word2VecGoogle}. During the training various parameters can be tuned which affects the performance, in terms of time and quality. The details on how the parameters are tuned and the effects of different values are presented in the evaluation section.% the Section \ref{eval}.

% TODO / NOTE: Removed
%In traditional recommendation process, the input data is composed of three base elements: user, item and rating. In most of the algorithms, these elements are represented by a \textit{user $x$ items} matrix, where the matrix entries indicate the ratings. For our venue recommendation problem, the ratings are assumed to be binary, such that the user is either checked in at a location(venue) or not. Then, each target user's past preferences can be represented as a list of items (the check-in venues). 

There are several similarities between the Word2Vec techniques and the recommendation process: First, the input data used in  Word2Vec techniques  is actually similar to what is used in the recommendation process. In the recommendation process a list of items that the user preferred/rated in the past are used and these lists can be divided into individual items. In other words, the sentences used in  Word2Vec can be mapped into past preferences of users in recommendation process and the words in  Word2Vec to individual items used in recommendation process. Second, the purpose of the  Word2Vec techniques and the recommendation process are similar.  Word2Vec model aims to predict the words based on the observed words, which can be mapped to predicting the items to be recommended based on already preferred/used items.

\begin{figure}
 \begin{subfigure}{0.5\linewidth}
   \centering
 	\includegraphics[width=0.95\linewidth]{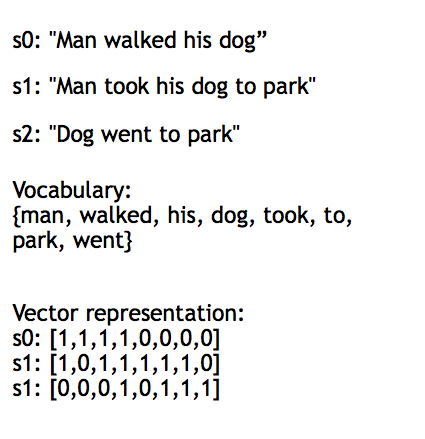}
  	\caption{Sentence-word data}
  	\label{sampleWords}
 \end{subfigure}%
\begin{subfigure}{.5\linewidth}
  \centering
  \includegraphics[width=.95\linewidth]{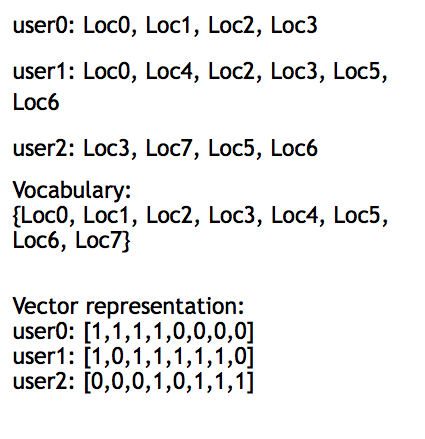}
     \caption{User-item data}
      \label{sampleItems}
  \end{subfigure}
\caption{Example data for text processing and recommendation systems}
 \label{wordsItems}
\end{figure}

In the Figure \ref{wordsItems}, the similarity of data used in text processing and recommendation systems are presented. On the left side of the figure, three sentences are presented together with the vocabulary list (dictionary). Similarly on the right side of the figure three users and their past preferences on locations that they have checked in are presented. Similar to sentence-word example, given in the figure \ref{sampleWords}, it is possible to create a list of locations (venues). Both of the examples can be represented as a vector. The vectors on the left present the existence of words in the related sentences and the vectors on the right show if the related user visited the locations (venues) or not.

In this work, inspiring from \cite{LeM14}, the item lists are used together with the users as the input to Word2Vec techniques, i.e. not only list of items, but list of user and the items preferred by this user is given as input to the Word2Vec techniques. As a result, the continuous vector representation of the items and the users are obtained, separately. These vector representations can be used to decide on which item is more similar to other item or which user is contextually closer to which items. These vectors and their similarities are used in the next step of our recommendation method.

\begin{figure*}
  \centering
    \begin{subfigure}{0.5\linewidth}
        \includegraphics[width=0.95\linewidth]{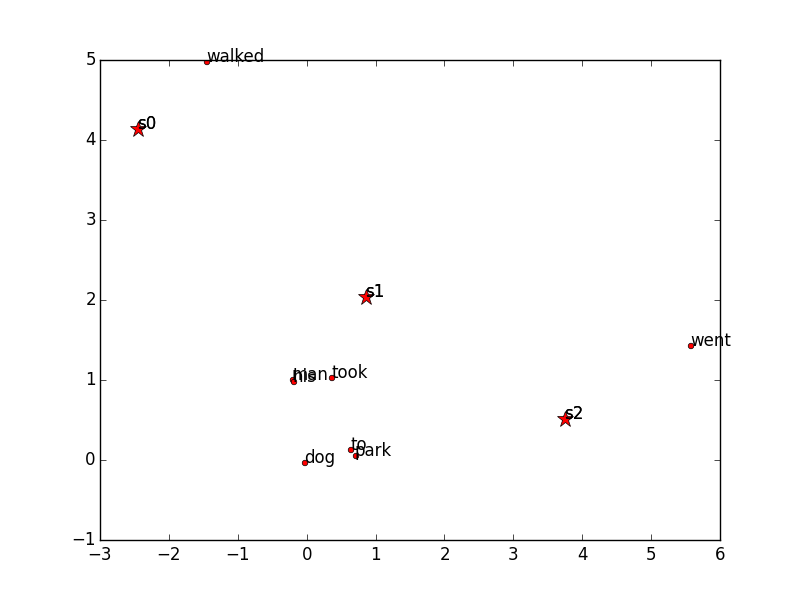}
        \caption{Sentence-word data}
        \label{contVecSampleWords}
 	\end{subfigure}%
    \begin{subfigure}{0.5\linewidth}
        \includegraphics[width=0.95\linewidth]{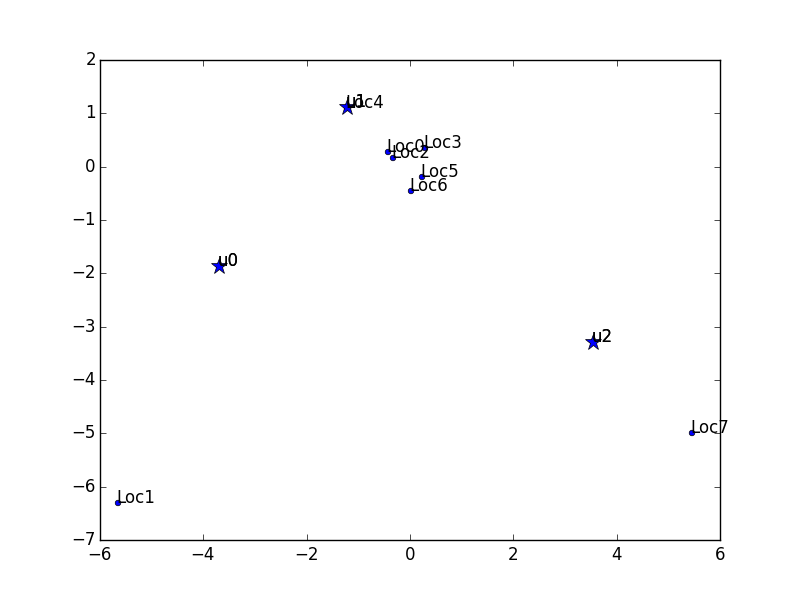}
        \caption{User-item data}
        \label{contVecSampleItems}
    \end{subfigure}
    \caption{Continuous vector representation of the example data for text processing and recommendation systems}
     \label{contVecWordsItems}
\end{figure*}

In the Figure \ref{contVecWordsItems}, the output of skip-gram technique on the input data given in the Figure \ref{wordsItems} is presented. To be able to plot the figure, the output continuous vector representation dimension is set to two. In the Figure \ref{contVecSampleWords}, the output for sentence-word data is shown. According to this figure, the relations among words and sentences are captured. For example, in the figure the word `walked' is closer (more related) to the sentence s0 and the word `went' is closer to the sentence s2;  these words are seen only in these sentences. Another example is the words `man' and `his' which are represented at the same position on the figure. This indicates that the technique is able to capture the relation between these words. In the Figure \ref{contVecSampleItems} the output for user-item data is shown. From the figure relations among users and items can be observed. For example, in the figure `Loc7' is represented closer to `u2' (user2); in the input data `Loc7' is visited only by `user2'. Another example is related to relations among locations;  in the figure `Loc0' and `Loc2' are closer to each other and in the input data these locations are always visited together.

\subsection{Recommendation using continuous vector representation} 
%The skip-gram model provides the word vectors where the words with similar meaning are located closer in the vector space \cite{LeM14}. In the recommendation case, instead of words, there are items and users. 
The output of Word2Vec techniques provide the continuous representation of the items and users in vector space where similar vectors are located closer to each other. In this report three different recommendation techniques that use the continuous vector representation of items and users are proposed:

\textbf{Recommendation by k-nearest items (KNI): } The k-nearest items (KNI) approach is inspired from traditional content-based filtering recommendation method. In content-based filtering method, features or tags that describe users and the items are used. For the recommendation the similarity between the target user and the items are calculated and the items that are more similar to the target user are listed as the recommendation. In KNI approach, instead of description of users and items, the continuous vector representations calculated in the previous step are used. For this purpose, the cosine similarity between the target user and item vectors are calculated and the most similar k items to the target user are found.  The collected top-k items are recommended to the target user. For example, given the vectors presented in the Figure \ref{contVecWordsItems}, assume that we want to make two location recommendations to the user u0. The most similar location vectors to the user vector belong to `Loc1' and `Loc2', so these locations are recommended to the target user.

%As it is possible to collect other users as one of the most similar items on the vector space, I included an additional processing step which removes unwanted items (i.e. other users) and re-collects new items to fill the top-k list that I want to collect. 

\textbf{Recommendation by N-nearest users (NN): } In recommendation by N-nearest users (NN) approach, the traditional user-based collaborative filtering method is applied on the continuous vector representations modeled in the previous step. In traditional user-based collaborative filtering, first the most similar users (neighbors) to the target user are selected, and then the items that are previously preferred by the neighbors are recommended to the target user. Similar to the traditional approach, in NN approach, first the top-N neighbors are decided using the similarity among the user vector representations. Then the items that are previously used/preferred by the top-N neighbors are collected. Summing up the votes of neighbors, the top-k items to recommend are decided. For example, using the previously presented example in the Figure \ref{wordsItems}, assume that we want to make two location recommendations to the user u0 by using a single neighbor. According to continuous vector representations (The Figure \ref{contVecWordsItems}), the most similar user of user0 is user1(u1) and it is selected as the neighbor. Two of the locations previously visited by user1, namely `Loc0', `Loc4', `Loc2', `Loc3', `Loc5', `Loc6', are chosen (randomly) and are recommended to the target user (u0).

\textbf{Recommendation by N-nearest users and k-nearest items (KIU): } This approach is a combination of the previous two approaches. In this approach, first, the top-N neighbors are found by using the vector representation of the users. Then the top-k items that are most similar to the combination of target user and the neighbors are found by using the vector representations calculated in the first step. The collected top-k items are recommended to the target user. For example, assume that we want to make two location recommendations to the user u0 by using a single neighbor. As explained in the previous method, the selected neighbor is u1 based on the continuous vector similarity. The two most similar location vectors to the user0's and user1's vector belong to `Loc2' and `Loc0'. These two locations are recommended to the target user, u0, by the KIU method.

\section{Evaluation}\label{eval}
In this work, the aim is to recommend k-many check-in venues to each user based on their past check-ins. For this purpose the Checkin2011 dataset \cite{GaoTL12} is used. The original dataset \footnote{http://www.public.asu.edu/~hgao16/dataset.html} is collected from Foursquare web-site in between January 2011 - December 2011 and contains 11326 users, 187218 locations, 1385223 check-ins and 47164 friendship links. However, in \cite{Ozsoy14} the researchers used a subset of this dataset by using the check-ins made in January as the training set and check-ins made in February as the test set, and named it as CheckinsJan. The CheckinsJan dataset contains 8308 users, 49521 locations and 86375 check-ins. In this work, the same sub-dataset, CheckinsJan dataset, is used for the experiments.

The performance of the methods is measured by Precision@k, Ndcg, Hitrate and Prediction Coverage metrics. While giving the evaluation results, for each user the performance metrics are calculated separately and then their averages are presented.

Precision@k measures the relevance of items on the output list. The Ndcg (Normalized discounted cumulative gain) metric decides the relevance of the listed items depending on their rank. Since these metrics are well-known in the literature and because of the limited space, we did not present how these metrics are calculated in this work. Hitrate measures the ratio of user who are given at least one true recommendation. In the Equation \ref{hitRate}, $m$ is one of the users, $M$ is the total set of users, $|M|$ is the size of the users and $HitRate_m$ indicates if there is a hit for the target user $m$ or not. $HitRate_m$ is equal to $1.0$ if there is at least one true recommendation for that user and $0.0$ otherwise. 

% The average precison@k value of a method can be high even though it is able to give recommendation just to a few users. For example, assume that I have two different recommendation methods, $RM1$ and $RM2$, two users $u$ and $v$ and the output list size is 3. Consider the case that $RM1$ gives 2 true recommendations to the user $u$ and no true recommendation to the user $v$, and $RM2$ gives 1 true recommendation to each of the users. Both methods' precision@k performance will be 0.33, on the average. However, $RM2$ can give true recommendations to both of the users, so I can say that it is better than the $RM1$. 

\begin{equation}\label{hitRate}
HitRate = \frac{\sum_{m \in M} HitRate_m}{|M|} 
\end{equation}

Prediction Coverage measures the ratio of the users who are given any recommendation, independent from being relevant or not. Some of the recommendation methods, e.g. collaborative filtering, may not be able to make any recommendation to some of the users who have not got any preference history or who have unique tastes. For example, in CheckinsJan dataset, the user81 visited ten venues which are never visited by any other user. Making recommendation to these kind of users, known as cold-start users in the literature, is a challenging task. In the recommendation systems literature, some of the methods may loose coverage in order to increase the accuracy \cite{Bellogin:2013:ECS:2414425.2414439}. \cite{Herlocker:2004:ECF:963770.963772} states that coverage and accuracy should be analyzed together.

The methods presented in this report use two parameters: Number of neighbors ($N$) and output list size ($k$). In  \cite{Ozsoy14} it is decided that best performing values are $N=30$ and $k=10$ for the CheckinsJan dataset. In this work, the values are directly used for the experiments. The upper-bound of the methods based on the decided parameters are also presented in \cite{Ozsoy14}: The upper-bound for Precision metric is found as $0.489$ and for the rest of the metrics they are found as $1.0$. 

% Training related results
Several different settings are evaluated when the input data is modeled using skip-gram and CBOW techniques. The parameters used during the training affects the performance in terms of time and quality \cite{Word2VecTutorial}. These parameters are based on the gensim toolbox implementation. In this work only the parameters that are set to a different value than the default are detailed; for the rest of the parameters one can refer to gensim web-page. The details of the parameters and how they are tuned are as follows:

\begin{itemize}
\item $min\_word\_count$: The technique ignores the items whose frequency is less than $min\_word\_count$ parameter. Its default value is 5. In natural language processing, the items (words) that are seen only few times can be considered as garbage or typo, however in the recommendation systems the data is very sparse and having items that are observed only few times is normal. In order not to lose any item that is not used frequently,this parameter is set to 1 during the experiments.
\item $size$: $size$ parameter represents the dimension of the feature vectors and its default value is 100. In \cite{Word2VecTutorial} it is stated that bigger $size$ value can lead more accurate model, but requires more data. The suggested values for $size$ parameter is in tens to hundreds \cite{Word2VecTutorial}. In our experiments this parameter is referred as $feature\_count(F)$ and is set to different values in the range of $[10,100]$ with 10 increments.
\item $window$: Window parameter assigns the maximum distance between the current and the predicted words and its default value is 5. \cite{Word2VecTutorial} states that it should be large enough to capture the semantic relationships among words. In our experiments this parameter is referred as  $context\_count(C)$ and is set to different values in the range of $[5,20]$ with 5 increments. However, during the experiments we observed that CBOW does not provide good performance results when small window size  is used. For CBOW we preferred to set the window size to the maximum length of input vectors.

\item $iter$:  $iter$ parameter represents the number of iterations on the input data and its default value is 1. In our experiments it is referred as  $epoch\_count(E)$ and is set to different values in the range of $[5,25]$ with 5 increments.
\end{itemize}

While iterating on the different ranges of one parameter, the other parameters are fixed to a constant value. For example while iterating on  $feature\_count$, the values of $context\_count$ and $epoch\_count$ are fixed. The constant values for the parameters are set to: $feature\_count(F)=100$, $epoch\_count(E)=25$, $context\_count(C)=20$ for skip-gram and $context\_count(C)=683$ for CBOW (The maximum length of input vector is 683 for CheckinsJan dataset). 

\begin{figure}
  \centering
  \includegraphics[width=0.95\linewidth]{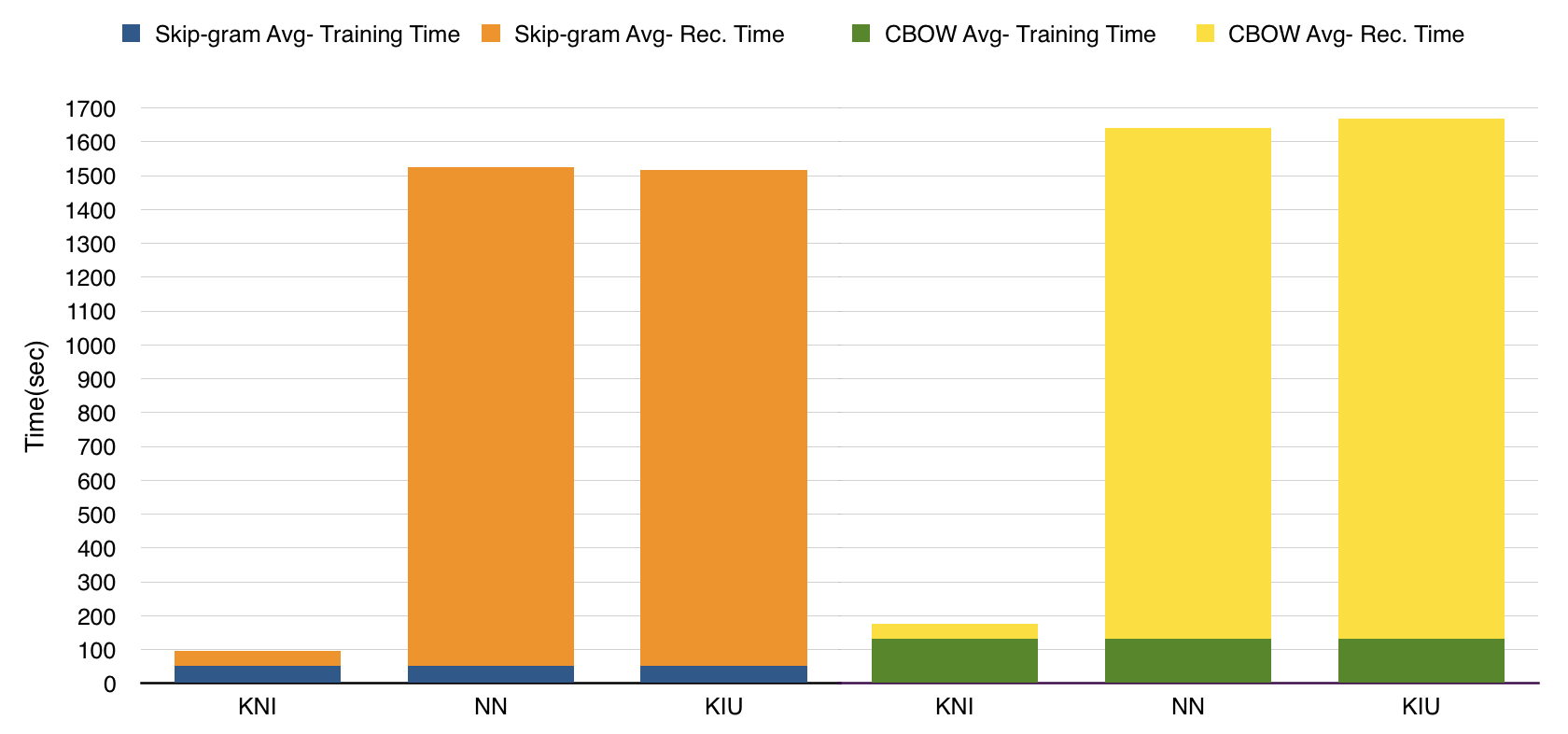}
  \caption{The time needed to model the input data and make recommendation}
\label{timeAll}
\end{figure}

% Time 
In this work, three different recommendation techniques that use the models trained by Word2Vec techniques are proposed. These techniques are `recommendation by k-nearest items (KNI)', `recommendation by N-nearest users (NN)' and `recommendation by N-nearest users and k-nearest items (KIU)'. The time spent to train the model by skip-gram or CBOW techniques and to make recommendation to all target users are presented on the Figure \ref{timeAll}. The presented time results are calculated by taking the average of time spent when combinations of $feature\_count$, $context\_count$ and $epoch\_count$ values are used.  According to the Figure \ref{timeAll}, training the model using the skip-gram technique spends less time than CBOW technique. Skip-gram technique spends about 50 seconds for training while CBOW spends about 150 seconds on the average. Among the proposed recommendation methods KNI method spends less time than other methods. This can be the result of the fact that this method directly uses the output of the model created by Word2Vec techniques without any further computations, such as finding neighbors. On the average KNI method spends less than 50 seconds to make recommendation to all target users, while both NN and KIU methods spend around 1500 seconds. These results indicate that the proposed methods spend less than 0.20 seconds for each target user in the recommendation step.

\begin{figure*}
  \centering
   \begin{subfigure}{0.33\textwidth}
       \includegraphics[width=0.95\linewidth]{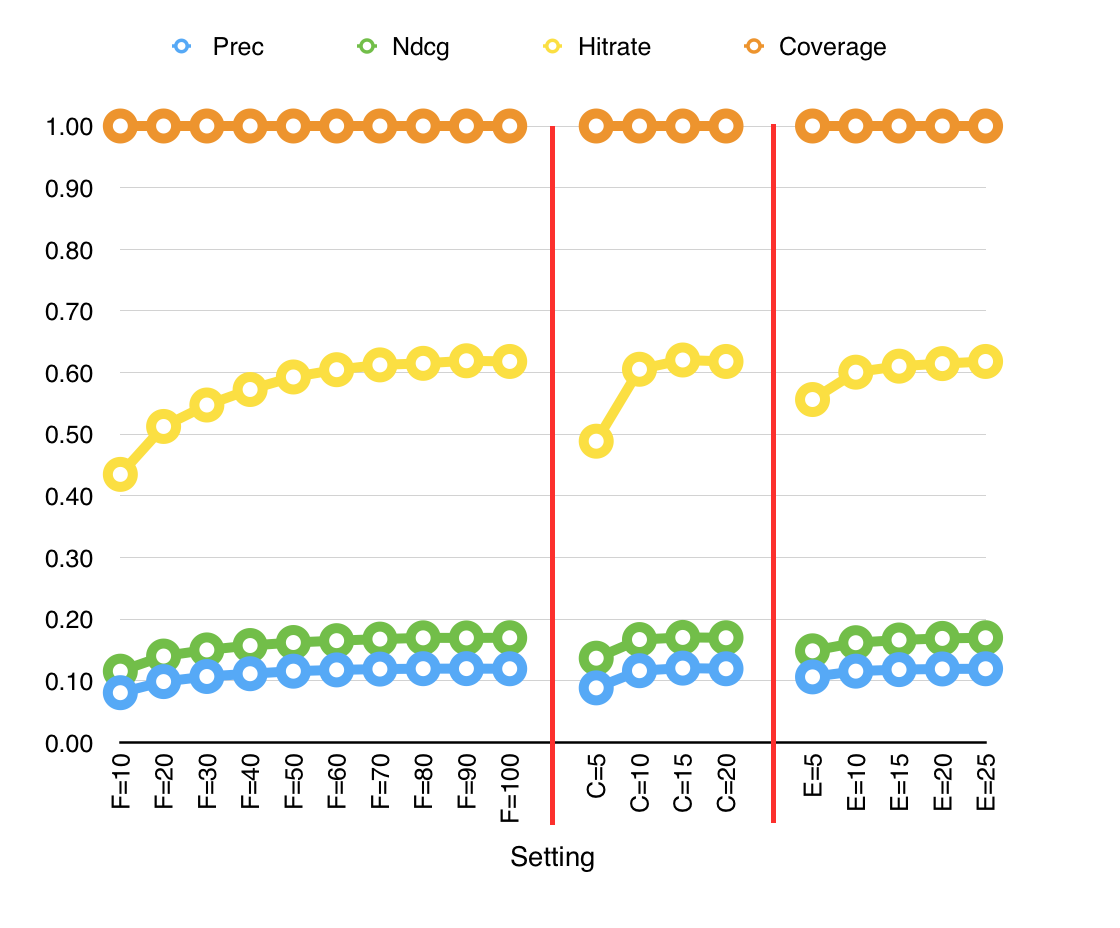}
		\caption{Performance results of KNI (Skip-gram)}
		\label{KNI_results}
 	\end{subfigure}%
 	   \begin{subfigure}{0.33\textwidth}
       \includegraphics[width=0.95\linewidth]{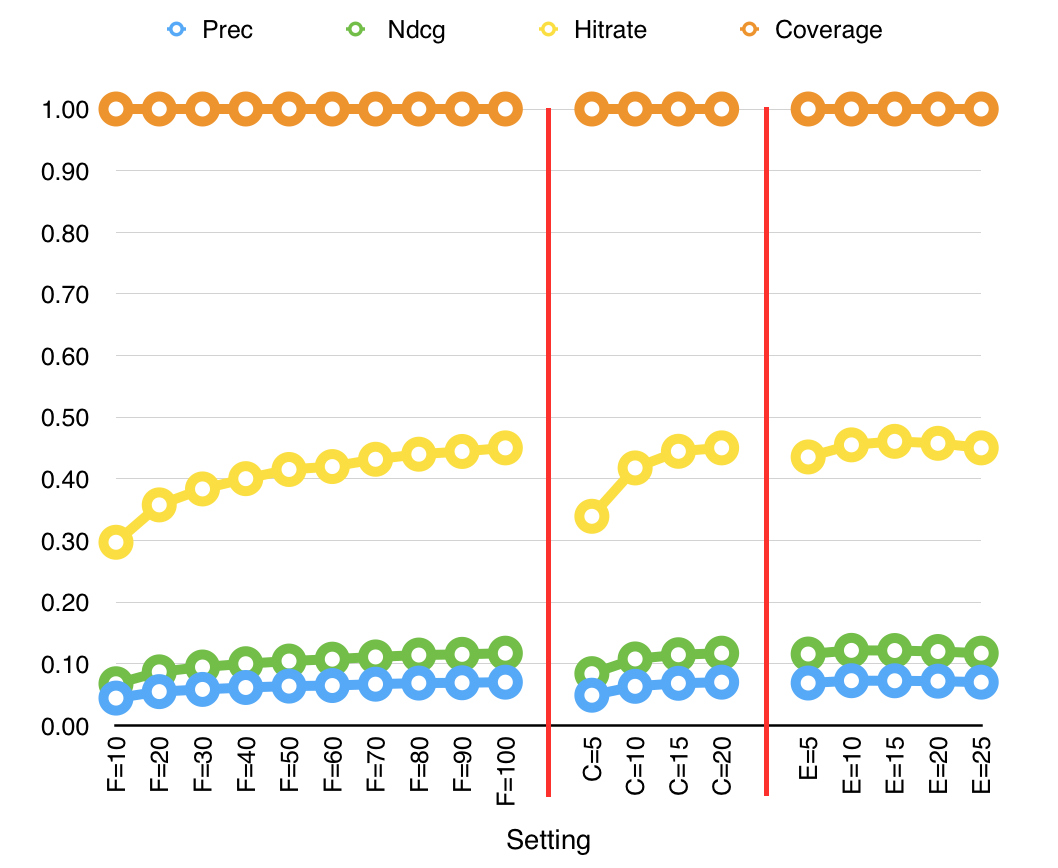}
		\caption{Performance results of NN (Skip-gram)}
		\label{NN_results}
 	\end{subfigure}%
 	   \begin{subfigure}{0.33\textwidth}
       \includegraphics[width=0.95\linewidth]{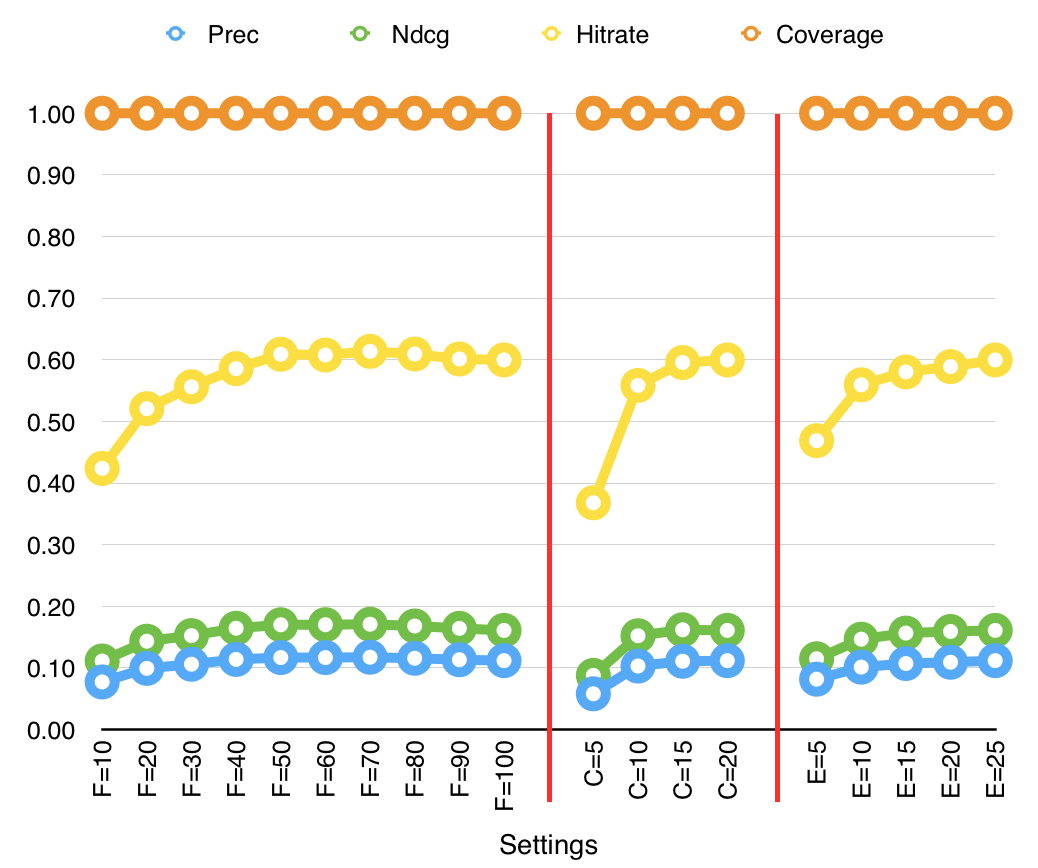}
		\caption{Performance results of KIU (Skip-gram)}
		\label{KIU_results}
 	\end{subfigure}
 	
    \begin{subfigure}{0.33\textwidth}
        \includegraphics[width=0.95\linewidth]{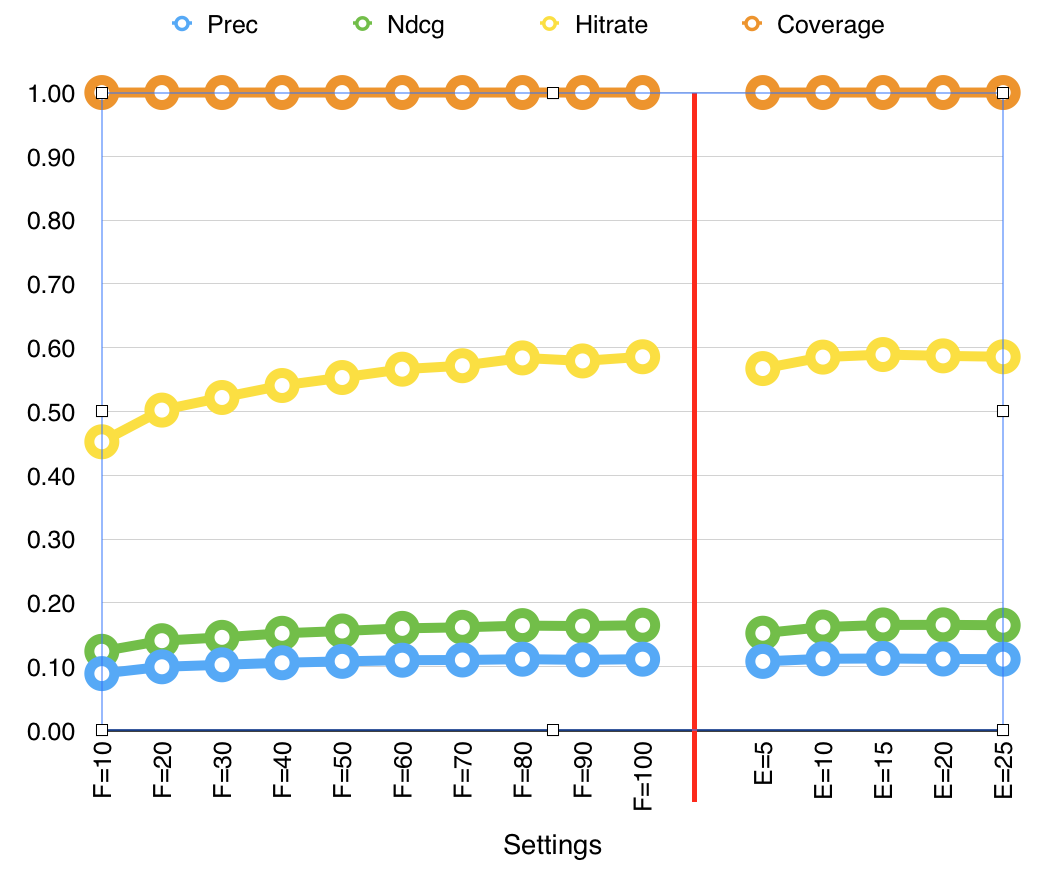}
		\caption{Performance results of KNI (CBOW)}
		\label{KNI_resultsCbow}
    \end{subfigure}%
    \begin{subfigure}{0.33\textwidth}
        \includegraphics[width=0.95\linewidth]{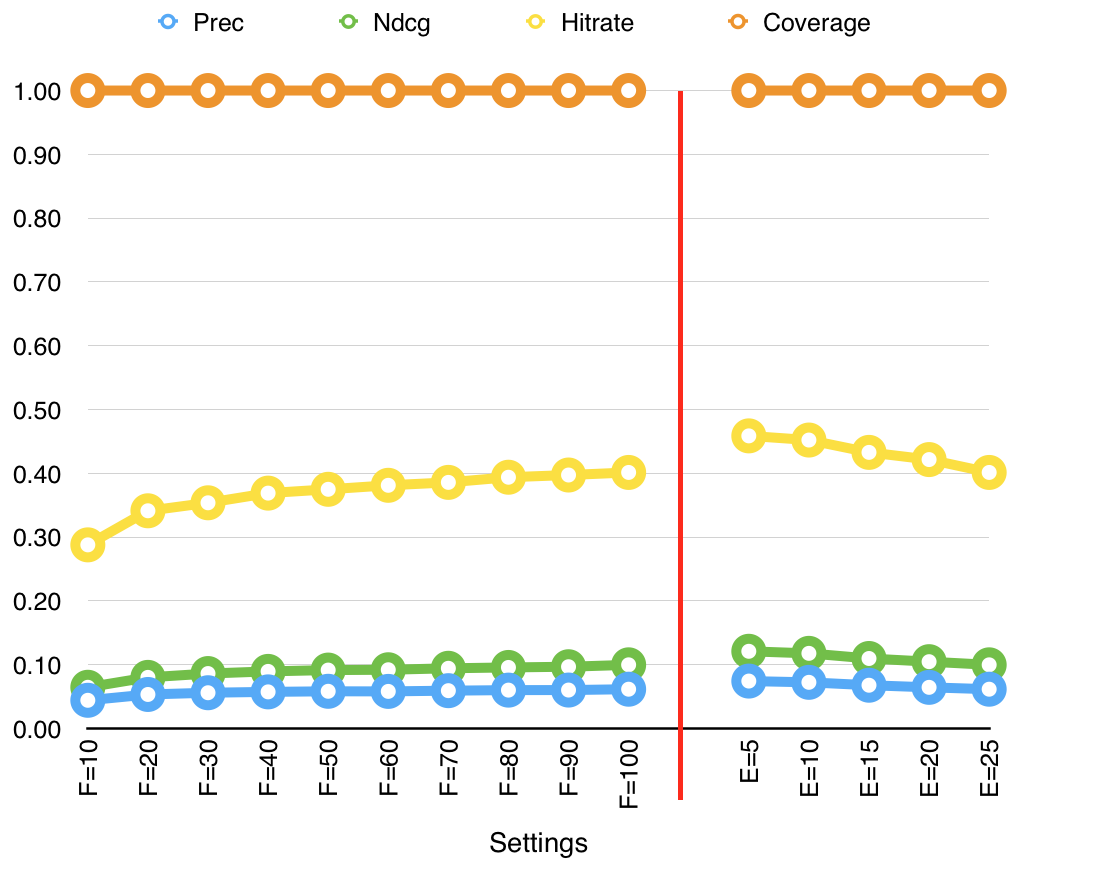}
		\caption{Performance results of NN (CBOW)}
		\label{NN_resultsCbow}
    \end{subfigure}%
    \begin{subfigure}{0.33\textwidth}
        \includegraphics[width=0.95\linewidth]{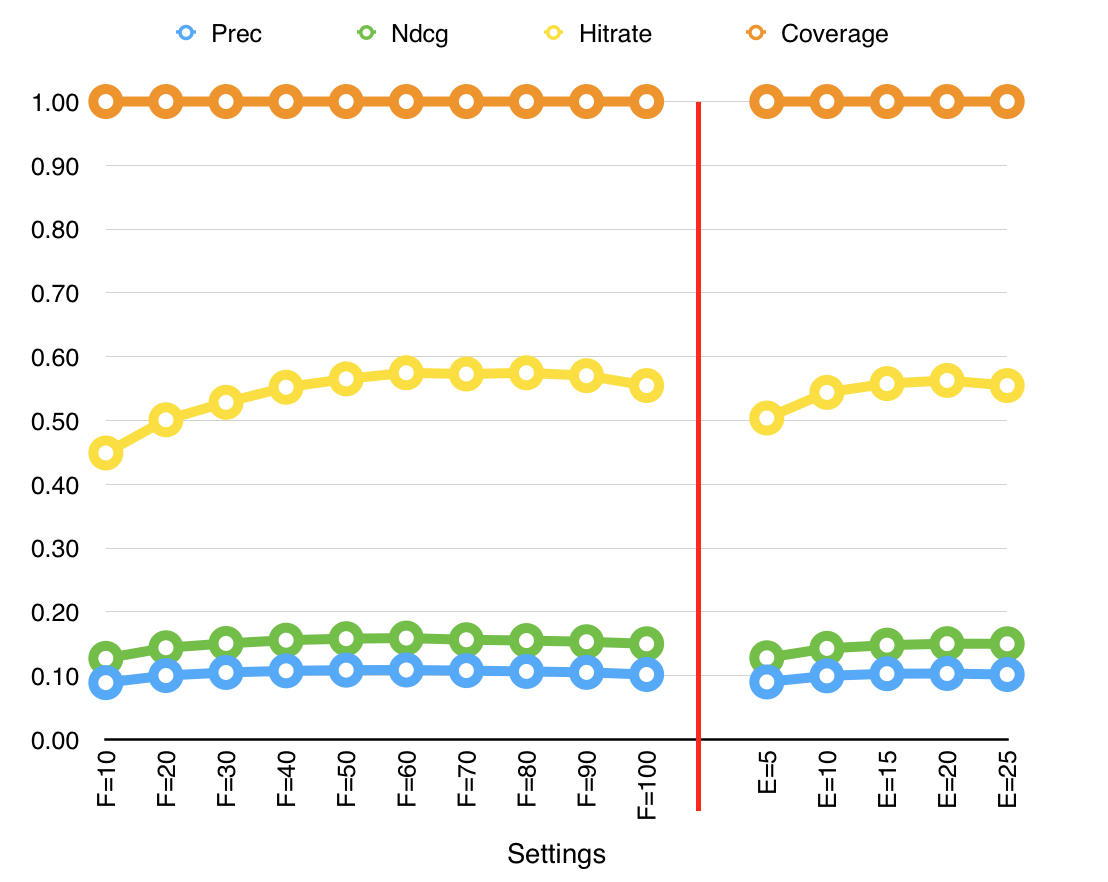}
		\caption{Performance results of KIU (CBOW)}
		\label{KIU_resultsCbow}
    \end{subfigure}
    \caption{Performance results of the proposed methods}
     \label{perfResults}
\end{figure*}

%Rec related results
The Figures \ref{KNI_results},  \ref{NN_results} and \ref{KIU_results} show the the recommendation performance of the proposed methods which use continuous vectors produced by skip-gram technique. For all of the methods, increasing the $feature\_count$, i.e. the size of the vector representation, improves the performance. The effect of the increase for this parameter is less obvious as it is set to higher values than 50. When $context\_count(C)=5$ the model is not able to capture the semantic similarity among the items and users. After setting this parameter to $C=10$ and higher, it performs better. For both $context\_count$ and $epoch\_count$, increase of the parameters slightly improves the performance. Similar observations can be made when CBOW is used as the first step to produce the continuous vector representations of the input data (The Figures \ref{KNI_resultsCbow},  \ref{NN_resultsCbow} and \ref{KIU_resultsCbow}). 

According to Figures \ref{KNI_results}-\ref{KIU_resultsCbow}, the comparison of the performance of the recommendation techniques indicates that the best performing method is KNI, followed by KIU. Both of these methods use the vector representations of the items and their similarities to the target users. Both KNI and KIU uses similarity to item vectors, however KIU additionally uses the neighbors of the target users, which are decided by user vector similarities. Compared to KNI, the use of neighbors in KIU does not provide high performance gain. This may indicate that use of user vector similarities for CheckinsJan dataset is not effective. This may root from the fact that the users could not be differentiated in the vector space efficiently since the number of users in the dataset is not very high, i.e. only 8307 users.

Comparison of the techniques that use CBOW to the techniques that use skip-gram shows that using the continuous vectors produced by skip-gram technique performs slightly better than using the ones produced by CBOW. This can be related to the characteristics of the input data. In the CheckinsJan dataset the check-ins in the predefined window do not necessarily related except they are visited by the same user. For example, the check-ins may be done in different time of the day (e.g. morning,  evening), but still listed in sequence; such that the related user did not visit any other venue in between. This may lead to poorer performance,  as CBOW uses window-sized data listed in sequence to decide on the relevance of the items (venues).

% TODO / NOTE  the time of check-ins info 

In the Table \ref{results}, the proposed KNI, NN and KIU methods are compared to the methods in the literature. The first method from the recommendation literature is traditional collaborative filtering method (CF-C) which uses the past preferences of the users and their similarities. We implemented this method to observe its performance. The second method is Random, which randomly chooses k-many items (venues) to recommend. We run Random method 10 times and presented the average of the results. 

The third and fourth methods use matrix factorization; the third one is SVD based recommendation, proposed by Sarwar et al. \cite{sarwar2000} and the fourth one CCD++ algorithm, is proposed by Yu et al. \cite{hfy12a}. In the original CCD++ method, rating prediction of  missing items are calculated, such that the method predicts the ratings of unseen items and recommends the items that have the highest predicted score. However, the problem proposed in this work does not aim to predict rating but aim to rank items. To solve this problem using CCD++, we adopted ideas from SVD based recommendation method \cite{sarwar2000} to CCD++ algorithm \cite{hfy12a}. In both of the methods, the input user-item matrix is decomposed into other matrices one of which represents the user-latent features relations (U). After reducing the dimension of the U matrix ($U_t$), recommendation is made by calculating the similarities among users-latent feature vectors, choosing the most similar users (neighbors) to the target user and listing the items which are previously used (e.g. venues that are previously visited) by the neighbors as recommendation. We used LIBPMF and sparsesvd libraries for  matrix factorization and implemented the rest of the method ourselves.

The fifth and sixth methods are from \cite{GaoTL12}, which uses a language model based method from natural language processing literature for recommendation. The method can combine past preferences and social ties of users. Two versions of the methods are proposed; one of which uses only the past preferences of the users (Gao-H) and the other uses the combination of past preferences and social ties (Gao-SH). We directly used the provided code in the authors' web-page. The last two methods are from \cite{Ozsoy14}, which are based on multi-objective optimization technique and combines past preferences of the users with other features, such as users' hometowns, friendship relations and their influence on each other. The methods are abbreviated as MO-CH when it uses past check-ins and hometowns of the users and MO-CFIH when it uses all of the above-mentioned features. For the experiments of MO based methods, the code provided by the authors is used.

\begin{table}
\renewcommand{\arraystretch}{1.1}
\caption{Comparison of methods}\label{results}
\centering
\begin{tabular}{c||c||c||c||c}
\hline
Method &  Precision &  Ndcg & HitRate &  Coverage \\
\hline
KNI(skip-gram) 		&	0.119	&	0.169	&	0.618	&	1.000 \\
NN(skip-gram) 			&	0.070	&	0.117	&	0.450	&	1.000 \\
KIU(skip-gram) 		&	0.112	&	0.161	&	0.599	&	1.000 \\
KNI(CBOW) 		&	0.112	&	0.165	&	0.586	&	1.000 \\
NN(CBOW) 			&	0.062	&	0.100	&	0.401	&	1.000 \\
KIU(CBOW) 		&	0.102	&	0.150	&	0.555	&	1.000\\
\hline
CF-C 		&   0.114 	&	0.242	&  0.621 	&	0.955 \\
Random		&	0.0001	&	0.0001	&	0.001	&	1.000 \\
SVD			&   0.058 	& 	0.104 	& 0.392 & 	1.000 \\
CCD++    &   0.073	& 	0.121 	& 0.461 & 	1.000 \\
Gao-H		&	0.174	&	0.299	&	0.696	&	0.952 \\
\hline
Gao-SH		&	0.167	&	0.295	&	0.721	&	0.992 \\
MO-CFIH 	&	0.105	&	0.218	&	0.596	&	0.999 \\
MO-CH		&	0.112	&	0.227	&	0.616	&	0.996\\		
\hline
\end{tabular}
\end{table}

In the Table \ref{results} first group of methods are the ones that are proposed in this paper, the second group are the ones that use a single feature (the past check-ins of the users) and the last group are the methods that combine multiple features to make recommendation. Among the proposed methods, the best performing method is KNI which uses continuous vector representation modeled by skip-gram technique. NN performs poorer than the others, which indicates that recommending items that are only used by neighbors is not able to capture the preference of the target user, and use of item similarity to the target user is more promising. The random method's poor performance show that making venue recommendation on CheckinsJan dataset is challenging, since there are many items to recommend, e.g. nearly 50000 venues. 

The NN, CF, SVD-based and CCD++ methods have similar steps: Finding neighbors and recommending venues that are visited by neighbors. Interestingly, the methods that learn low-dimensional vector space representation of the input data (NN, SVD-based and CCD++) are worse than the method that directly use user similarity (CF). This may be result of the sparsity of the CheckinsJan data, such that NN, SVD-based and CCD++ methods are not able to model the user relations as well as modeling the user-item and  item-item relations. 

Among the listed methods, the best performing methods are Gao-H and Gao-SH, which use language model technique from natural language processing. This result shows that use of methods from natural language processing can be effective for making recommendations. Even though GAo-H and Gao-SH are better at accuracy related metrics, they are not able to make predictions for some of the users. For example, having 0.952 prediction coverage ratio, Gao-H cannot make recommendation to about 350 of the users.  Some of the methods in the literature may not make any recommendation to more challenging users, such as users who have not got any preference history or who have unique tastes. Making recommendation to these challenging users may lead to reduced accuracy results, since it is harder to make true recommendations for them. In terms of prediction coverage, the methods proposed in this paper and matrix factorization methods are the better than the others. Overall, results show that use of Word2Vec techniques produce comparable performance to the methods in the literature and they are promising for making recommendations.

% TODO Write sth about third group
%The method that combine multiple features generally perform better in term of Ndcg while preserving the other metrics. This  indicates that combining multiple features help the methods to produce not better recommendations but better output lists

\section{Conclusion}\label{conclusion}
Recommendation systems predict the future preferences of users based on their previous interactions with the items. In the literature, there are many different techniques to make recommendations, e.g. neighborhood based, machine-learning based and matrix-factorization based methods. In this work, we employed Word2Vec's skip-gram and CBOW techniques to make next check-in venue (location) recommendations on Location Based Social Networks (LBSNs). We proposed several different techniques that combines \lword{Word2Vec} techniques and the well-known recommendation methods, namely content based and collaborative filtering based methods. Unlike the previous works that use Word2Vec for recommendation (\hspace{1sp}\cite{ShinCL14}, \cite{MustoSGL15}), we used a non-textual feature, namely the past check-ins of the users, to make recommendations. For the evaluation a Foursquare check-in dataset, which is already used in previous works (\hspace{1sp}\cite{GaoTL12}, \cite{Ozsoy14}) is employed. Also comparisons to methods from the recommendation systems literature are presented. The results show that use of techniques from natural language processing is effective and use of Word2Vec techniques is promising for making recommendations.

In the future, more experiments will be done using the other datasets used in the recommendation literature to  observe the effectiveness of the proposed methods. Also, we will integrate multiple features while using continuous vector space representations, as previous works showed that combining multiple features increases the recommendation performance.

% conference papers do not normally have an appendix

% use section* for acknowledgement
%\section*{Acknowledgment}
%The authors would like to thank...

% trigger a \newpage just before the given reference
% number - used to balance the columns on the last page
% adjust value as needed - may need to be readjusted if
% the document is modified later
%\IEEEtriggeratref{8}
% The "triggered" command can be changed if desired:
%\IEEEtriggercmd{\enlargethispage{-5in}}

% references section

% can use a bibliography generated by BibTeX as a .bbl file
% BibTeX documentation can be easily obtained at:
% http://www.ctan.org/tex-archive/biblio/bibtex/contrib/doc/
% The IEEEtran BibTeX style support page is at:
% http://www.michaelshell.org/tex/ieeetran/bibtex/
%\bibliographystyle{IEEEtranS}
% argument is your BibTeX string definitions and bibliography database(s)
%\bibliography{IEEEabrv,refs}
%
% <OR> manually copy in the resultant .bbl file
% set second argument of \begin to the number of references
% (used to reserve space for the reference number labels box)

%\begin{thebibliography}{1}

%\bibitem{IEEEhowto:kopka}
%H.~Kopka and P.~W. Daly, \emph{A Guide to \LaTeX}, 3rd~ed.\hskip 1em plus
%  0.5em minus 0.4em\relax Harlow, England: Addison-Wesley, 1999.

%\end{thebibliography}

% that's all folks
\end{document}